\documentclass[12pt]{article}

\usepackage{amssymb,amsmath,amsfonts,latexsym,graphicx}
\usepackage[numbers]{natbib}
\usepackage{hyperref}
\usepackage{float} 

\hypersetup{
    colorlinks=true,    
    linkcolor=black,    
    citecolor=blue,    
    filecolor=black,     
    urlcolor=blue      
}

\begin{document}

\begin{center}
\Large \bf CapsoNet: A CNN-Transformer Ensemble for Multi-Class Abnormality Detection in Video Capsule Endoscopy \rm

\vspace{1cm}


\large Arnav Samal$\,^\star$, \large  Ranya Batsyas$\,^\dagger$

\vspace{0.5cm}

\normalsize

$^\star$ Department of Computer Science and Engineering, National Institute of Technology Rourkela, India \\
$^\dagger$ Department of Artificial Intelligence and Data Sciences, Indira Gandhi Delhi Technical University for Women, Delhi, India

\vspace{5mm}


Email: {\tt 122cs0107@nitrkl.ac.in, ranya102btcsai22@igdtuw.ac.in}

\vspace{1cm}

\end{center}

\abstract{We present CapsoNet, a deep learning framework developed for the Capsule Vision 2024 Challenge, designed to perform multi-class abnormality classification in video capsule endoscopy (VCE) frames. CapsoNet leverages an ensemble of convolutional neural networks (CNNs) and transformer-based architectures to capture both local and global visual features. The model was trained and evaluated on a dataset of over 50,000 annotated frames spanning ten abnormality classes, sourced from three public and one private dataset. To address the challenge of class imbalance, we employed focal loss, weighted random sampling, and extensive data augmentation strategies. All models were fully fine-tuned to maximize performance within the ensemble. CapsoNet achieved a balanced accuracy of 86.34\% and a mean AUC-ROC of 0.9908 on the official validation set, securing Team Seq2Cure 5th place in the competition. Our implementation is available at \href{https://github.com/arnavs04/capsule-vision-2024}{this https URL}.
}

\section{Introduction}

\subsection{Background \& Challenges}

Gastrointestinal (GI) endoscopy is a critical diagnostic tool for identifying and managing various GI disorders, including celiac disease, gastrointestinal bleeding (GIB), esophagitis, and malignancies. Traditional endoscopy, while effective, is invasive and often requires sedation, which can be uncomfortable for patients.

Video Capsule Endoscopy (VCE), introduced in 2001, offers a non-invasive alternative. Patients swallow a small, wireless capsule that captures thousands of images as it moves through the GI tract, allowing for detailed visualization of small intestinal disorders like bleeding, erosions, and ulcers without the need for sedation.

Despite its advantages, VCE presents challenges, primarily due to the vast volume of data generated. Each procedure produces thousands of images, requiring significant time and expertise to analyze. The rapid movement of the capsule often leads to complex and stochastic imagery, increasing the likelihood of missed diagnoses. These challenges highlight the need for automated analysis tools to support efficient and accurate interpretation of VCE data.

To address these challenges, the Capsule Vision 2024 Challenge \cite{handa2024capsulevision2024challenge} was launched. The aim of the challenge is to provide a platform for the development, testing, and evaluation of AI models for automatic classification of abnormalities captured in VCE frames.

\subsection{Previous Works}
Recent studies have explored the application of artificial intelligence and machine learning tools (AIMLT) to enhance the efficiency and accuracy of VCE image analysis. Various research groups have employed convolutional neural networks (CNNs) to tackle challenges in VCE. \citet{xie2020} trained a CNN on nearly 3,000 capsule studies, achieving a significant increase in the detection rate of small bowel pathologic findings while reducing reading time by 89.3\% compared to conventional human reading, though the study faced limitations due to the heterogeneity of VCE readers involved.

Building on this foundation, \citet{afonso2019} developed a CNN specifically focused on identifying ulcers and erosions in the small bowel, achieving impressive results with a sensitivity of 90.8\%, specificity of 97.1\%, and an overall accuracy of 95.6\%. Further advancing this line of research, \citet{ferreira2021} demonstrated similar success in ulcer detection, achieving an accuracy of 92.4\%.

\subsection{Limitations and Proposed Solutions}
Previous research in VCE analysis has encountered several key challenges. The need for extensive pre-processing and labeling of VCE images to create reliable training datasets remains resource-intensive. Additionally, traditional CNN-based approaches struggle with capturing long-range dependencies due to fixed receptive fields, limiting their ability to detect subtle abnormalities in VCE frames. The high computational requirements of advanced machine learning models further hinder their use in clinical settings.

Our project addresses these challenges by implementing a multi-model ensemble approach that combines CNN and transformer architectures. This approach enhances the model’s ability to capture global dependencies and contextual information, leveraging the strengths of both architectures to improve classification accuracy and robustness.

\section{Methodology}\label{sec2}

\subsection{Dataset}
The training \cite{Handa2024training} and testing \cite{Handa2024testing} datasets were developed using a combination of three publicly available sources—SEE-AI project, KID, and Kvasir-Capsule—along with a private dataset from AIIMS. In total, the training set \cite{Handa2024training} comprised 37,607 frames, and the validation set included 16,132 frames, each annotated with one of 10 abnormality classes: angioectasia, bleeding, erosion, erythema, foreign body, lymphangiectasia, polyp, ulcer, worms, and normal. 

\begin{figure}[H]
    \centering
    \includegraphics[width=0.375\linewidth]{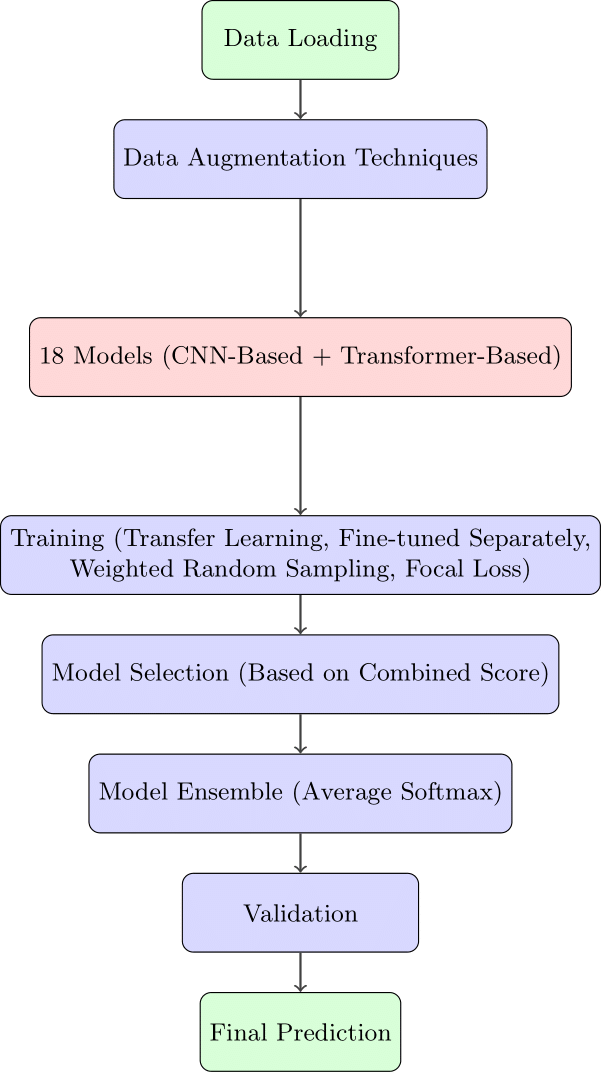}
    \caption{Block diagram of the developed pipeline.}
    \label{fig:enter-label}
\end{figure}

\subsection{Data Preprocessing}
To enhance model generalization, a comprehensive data augmentation pipeline was implemented, artificially expanding the training dataset \cite{Handa2024training} through several spatial and intensity transformations:

\subsubsection{Spatial transformations}
\begin{itemize}
\item \textbf{Resize \& Flip:} Images were resized to (224, 224) pixels, with a 50\% chance of horizontal flip and a 30\% chance of vertical flip.
\item \textbf{Rotation \& Affine:} Random rotations up to ±15°, with affine transformations including translation (up to 10\%) and scaling (90-110\%).
\item \textbf{Perspective Distortion:} A 50\% chance of applying perspective distortion to simulate variations in camera angles.
\end{itemize}
\subsubsection{Intensity transformations}
\begin{itemize}
\item \textbf{Color Jitter \& Normalization:} Adjusted brightness, contrast, saturation, and hue, followed by normalization (mean and std [0.5, 0.5, 0.5]).
\item \textbf{Random Erasing \& Gaussian Blur:} A 20\% chance of erasing parts of the image and a 30\% chance of applying Gaussian blur for robustness.
\end{itemize}

\subsection{Model Architecture}

This project employs a multi-model ensemble approach that integrates both traditional Convolutional Neural Network (CNN) architectures and transformer-based models for effective video capsule endoscopy (VCE) frame classification. This ensemble method aims to leverage the strengths of each model type, enhancing overall classification performance.

\subsection*{Traditional CNN Models}

The following CNN architectures are utilized due to their proven efficacy in image classification tasks:

\begin{itemize} 
\item \textbf{EfficientNet:} Optimizes the network architecture by balancing depth, width, and resolution through a compound scaling method \cite{tan2019efficientnet}.
\item \textbf{ResNet:} Known for its deep residual learning framework, which mitigates the vanishing gradient problem, enabling the training of very deep networks \cite{he2016deep}.
\item \textbf{MobileNetV3:} Designed for mobile and edge devices, focusing on high efficiency and performance with limited computational resources \cite{howard2019searching}.
\item \textbf{RegNet:} A family of models emphasizing design simplicity and efficiency with a straightforward, regularized architecture \cite{radosavovic2020designing}.
\item \textbf{DenseNet:} Utilizes dense connectivity patterns to improve feature propagation and reduce the number of parameters \cite{huang2017densely}.
\item \textbf{InceptionV4:} Enhances classification performance through a multi-path architecture incorporating various convolutional operations in parallel \cite{szegedy2017inceptionv4}.
\item \textbf{ResNeXt:} Combines the strengths of ResNet with the cardinality concept to create models that achieve improved accuracy and efficiency \cite{xie2017aggregated}.
\item \textbf{WideResNet:} Focuses on widening layers rather than deepening them to enhance performance with fewer layers \cite{zagoruyko2016wide}.
\item \textbf{MNASNet:} Optimized for mobile devices, using a reinforcement learning approach for architectural search \cite{tan2019mnasnet}.
\item \textbf{SEResNet50:} Integrates Squeeze-and-Excitation blocks into the ResNet architecture, improving representational capacity \cite{hu2018squeeze}.
\item \textbf{ConvNeXt:} A modernized CNN architecture that adopts concepts from transformers to enhance performance \cite{liu2022convnext}.
\end{itemize}

\subsection*{Transformer-based Models}

The following transformer architectures are employed for their advanced capabilities in handling image data:

\begin{itemize} 
\item \textbf{Vision Transformer (ViT):} Processes images as sequences, leveraging the self-attention mechanism to capture global dependencies effectively \cite{dosovitskiy2020image}.
\item \textbf{Swin Transformer:} Introduces hierarchical feature maps and shifted windows to improve computational efficiency and performance in vision tasks \cite{liu2021swin}.
\item \textbf{DeiT (Data-efficient Image Transformers):} Designed to work effectively with fewer data samples while maintaining competitive performance \cite{touvron2021training}.
\item \textbf{BEiT (Bidirectional Encoder Representation from Image Transformers):} Integrates a vision transformer with masked image modeling to provide rich contextual representations \cite{bao2021beit}.
\item \textbf{CaiT (Class-Attention in Image Transformers):} Enhances the representation of image classes using attention mechanisms within the transformer framework \cite{touvron2021training}.
\item \textbf{TwinsSVT (Spatially Separable Vision Transformer):} Utilizes spatial separability to improve efficiency while retaining high performance \cite{zhang2021twins}.
\item \textbf{EfficientFormer:} Combines efficiency and performance by focusing on the optimization of transformer structures for image classification tasks \cite{lin2021efficientformer}.
\end{itemize}

The model parameters are available at \href{https://www.kaggle.com/models/arnavs19/capsule-vision-2024-models}{this https URL}.
  
\subsection{Training}
To optimize performance and mitigate overfitting, the following techniques and configurations were applied:

\subsubsection{Transfer Learning \& Fine-tuning}
Pre-trained models were used for transfer learning, with \textbf{full model fine-tuning} performed on the video capsule endoscopy (VCE) dataset. All layers were fine-tuned to adapt the networks to domain-specific features, ensuring effective learning from the VCE data.

\subsubsection{Early Stopping}
\textbf{Early stopping} \cite{prechelt1998automatic} was employed with a patience of 5 epochs, halting training if validation performance did not improve for 5 consecutive epochs. This approach prevented overfitting and unnecessary training cycles.

\subsubsection{Optimizer \& Training Configuration}
The models were optimized using the \textbf{AdamW optimizer} \cite{loshchilov2019decoupled}, with the following settings:
\begin{itemize}
  \item \textbf{Learning rate:} 1e-4
  \item \textbf{Weight decay:} 0.05
\end{itemize}
\noindent The training process involved:
\begin{itemize}
  \item \textbf{Batch size:} 32
  \item \textbf{Epochs:} 20
  \item \textbf{Hardware:} 4 Nvidia Tesla P100 GPUs (16 GB memory each)
  \item \textbf{Training time:} 15-16 hours
\end{itemize}

\subsection{Class Imbalance Mitigation}
To address class imbalance, the following techniques were applied:

\subsubsection{Weighted Random Sampling}
\textbf{Weighted Random Sampling} \cite{byrd2019effect} was used to ensure more frequent sampling of underrepresented classes, reducing bias toward overrepresented classes. The sampling weight \( w_i \) for class \( i \) was defined as:

\[
w_i = \frac{1}{N_i}
\]

where \( N_i \) is the number of samples in class \( i \). This ensures that the probability of selecting samples from underrepresented classes is higher, promoting a balanced representation of all classes during training.

\subsubsection{Focal Loss}
\textbf{Focal Loss} \cite{lin2017focal} was applied to prioritize difficult-to-classify examples, effectively down-weighting well-classified samples and focusing more on hard examples. The focal loss for a binary classification task is defined as:

\[
\mathcal{L}_{\text{FL}}(p_t) = -\alpha_t (1 - p_t)^\gamma \log(p_t)
\]

where:
\begin{itemize}
  \item \( p_t \) is the predicted probability for the true class.
  \item \( \alpha_t \) is a balancing factor for the class (used to address class imbalance).
  \item \( \gamma \) is the focusing parameter, which adjusts the rate at which easy examples are down-weighted. A common value is \( \gamma = 2 \).
\end{itemize}

This loss function ensures that harder examples receive more focus by reducing the contribution of well-classified samples, effectively handling the class imbalance problem.

\subsection{Model Selection \& Checkpointing}
Model selection was based on a balanced evaluation metric that combined both \textbf{Balanced Accuracy} and \textbf{Mean AUC Score}, ensuring robust performance across all abnormality classes. This approach mitigated overfitting to dominant classes and enhanced the model's ability to detect rare abnormalities.

\noindent The best models were saved using a \textbf{Combined Score}, defined as:
\[
\text{Combined Score} = \frac{\text{Balanced Accuracy} + \text{Mean AUC Score}}{2}
\]
A tolerance of 1e-4 was used to track performance improvements, ensuring that only models demonstrating substantial gains were checkpointed, thus preserving the most optimal versions.

\subsection{Model Ensembling}
To implement the ensemble inference for the validation and test dataset \cite{Handa2024testing}, we used the following procedure, which leverages multiple models to generate predictions through softmax probability averaging. The pseudocode for the ensemble inference is as follows:

\begin{verbatim}
function ensemble(models, dataloader, device):
    all_predictions = []
    all_image_paths = []

    # Set models to evaluation mode
    for model in models:
        model.eval()

    # Inference with ensemble
    for (X, image_paths) in dataloader:
        X = X.to(device)
        
        # Compute average softmax predictions across models
        ensemble_preds = mean(softmax(model(X)) for model in models)
        
        all_predictions.append(ensemble_preds.cpu())
        all_image_paths.extend(basename(path) for path in image_paths)

    predictions = concatenate(all_predictions)
    return predictions, all_image_paths
\end{verbatim}

This method ensures that predictions from each model are combined, enhancing the overall robustness of the inference process through softmax probability averaging.

\section{Results}\label{sec3}
The evaluation of the ensemble model on the validation dataset demonstrates strong overall performance in classifying video capsule endoscopy (VCE) images. Our team’s submission ranked 5th, highlighting the competitive performance of the model.The model achieved a balanced accuracy of \(0.8634\) and a mean AUC-ROC of \(0.9908\), indicating its effectiveness in distinguishing various abnormalities. Notably, the model excelled in identifying worms, normal instances, and ulcers. However, it revealed areas for improvement, particularly in classifying erythema and polyps, which exhibited lower F1 scores. For detailed performance metrics, including class-wise precision, recall, specificity, and AUC-ROC scores, refer to \href{https://github.com/arnavs04/capsule-vision-2024/blob/main/submission/metrics_report.json}{here}.

\subsection{Achieved Results on the Validation Dataset}\label{subsec1}
The performance of the ensemble model was assessed on the validation dataset, yielding the key metrics outlined in Table \ref{tab:class_results}.

\begin{table}[h]
    \centering
    \resizebox{0.75\textwidth}{!}{ 
        \begin{tabular}{|p{2.75cm}|c|c|c|c|} 
            \hline
            \textbf{Class} & \textbf{Precision} & \textbf{Recall} & \textbf{F1-Score} & \textbf{Specificity} \\ 
            \hline
            Angioectasia & 0.87 & 0.84 & 0.85 & 0.99 \\ 
            Bleeding & 0.87 & 0.85 & 0.86 & 0.99 \\ 
            Erosion & 0.80 & 0.80 & 0.80 & 0.98 \\ 
            Erythema & 0.69 & 0.64 & 0.66 & 0.99 \\ 
            Foreign Body & 0.86 & 0.92 & 0.89 & 0.99 \\ 
            Lymphangiect. & 0.85 & 0.93 & 0.89 & 0.99 \\ 
            Normal & 0.98 & 0.98 & 0.98 & 0.94 \\ 
            Polyp & 0.78 & 0.73 & 0.75 & 0.99 \\ 
            Ulcer & 0.98 & 0.93 & 0.96 & 1.00 \\ 
            Worms & 0.99 & 1.00 & 0.99 & 1.00 \\ 
            \hline
            \textbf{Macro Avg} & 0.87 & 0.86 & 0.86 & 0.99 \\ 
            \hline
        \end{tabular}
    } 
    \caption{Performance Metrics for Each Class}
    \label{tab:class_results}
\end{table}

\vspace{0.25cm}

\begin{table}[htbp]
    \centering
    \renewcommand{\arraystretch}{1.5} 
    \resizebox{\textwidth}{!}{
        \begin{tabular}{ccccccc}
            \hline
            \textbf{Method}              & \textbf{\begin{tabular}[c]{@{}c@{}}Avg. \\ Acc.\end{tabular}} & \textbf{\begin{tabular}[c]{@{}c@{}}Avg. \\ Prec.\end{tabular}} & \textbf{\begin{tabular}[c]{@{}c@{}}Avg. \\ AUC\end{tabular}} & \textbf{\begin{tabular}[c]{@{}c@{}}Avg. \\ Rec.\end{tabular}} & \textbf{\begin{tabular}[c]{@{}c@{}}Avg. \\ F1\end{tabular}} & \textbf{\begin{tabular}[c]{@{}c@{}}Bal. \\ Acc.\end{tabular}} \\ \hline
            \textbf{SVM (baseline)}    & 0.82         & 0.83           & 0.94         & 0.41         & 0.49          & 0.41 \\ \hline
            \textbf{ResNet50 (baseline)} & 0.76         & 0.60           & 0.87         & 0.32         & 0.37           & 0.32 \\ \hline
            \textbf{VGG16 (baseline)} & 0.69         & 0.52           & 0.92         & 0.54         & 0.48           & 0.54 \\ \hline
            \textbf{Custom CNN (baseline)} & 0.46         & 0.10           & 0.31         & 0.10         & 0.09           & 0.10 \\ \hline
            \textbf{Ensemble Model (Ours)} & \textbf{0.946} & \textbf{0.905} & \textbf{0.990} & \textbf{0.863} & \textbf{0.864} & \textbf{0.863} \\ \hline
        \end{tabular}
    }
    \caption{Validation results and comparison to baseline methods from the Capsule Vision 2024 challenge.}
    \label{tab:results}
\end{table}

\section{Discussion}\label{sec4}
Based on our previous studies on medical image classification using CNN, we introduced a methodology that implements a multi-model ensemble approach for video capsule endoscopy frame classification. This approach combines the strengths of various CNN and transformer architectures to enhance the classification accuracy of VCE images.

\subsection{Rationale for using multi-model ensemble approach}
The decision to implement a multi-model ensemble approach for video capsule endoscopy (VCE) frame classification was driven by several factors. This approach enhances accuracy and robustness by leveraging various CNN and transformer architectures. It improves classification performance by combining diverse model capabilities to capture a broader range of features and reduce errors through averaged predictions, thereby increasing output reliability.

\subsection{Methodological Considerations}
Training models using GPUs incurred significant computational costs, presenting barriers to efficient model development. Although the models were pretrained, they were often trained on large datasets that contained only a few similar samples to the target dataset, limiting the effectiveness of transfer learning. 

Advanced training techniques like Knowledge Distillation, K-Fold Cross-Validation, and Self-Supervised Pre-training methods such as Masked Auto-Encoding and Contrastive Learning can enhance model performance and generalization.

\subsection{Challenges and Future Directions}
Despite promising results, challenges persist in applying AI in VCE. The dependence on extensive human annotation for training datasets is labor-intensive, potentially hindering clinical integration. We aim to develop a user-friendly annotation tool to enhance efficiency in dataset preparation and improve supervised CNN accuracy. Additionally, addressing the “black box” nature of AI remains essential; incorporating visual representations of areas of interest, like Grad-CAMs, can enhance model explainability. Finally, the imbalanced nature of VCE datasets calls for innovative approaches, such as few-shot learning, to mitigate the impact of limited pathology examples.

\section{Conclusion}\label{sec5}
In conclusion, our study highlights the potential of employing multiple CNNs to enhance the accuracy of lesion detection in the gastrointestinal tract using video capsule endoscopy, despite challenges associated with data limitations. The implementation of our AI approach offers promising pathways for improving diagnostic accuracy while alleviating burdens on medical professionals. Future research should focus on validating our approach across diverse datasets and refining our annotation tools for broader clinical application.

\section{Acknowledgments}\label{sec6}
As participants in the Capsule Vision 2024 Challenge, we fully comply with the competition's rules as outlined in \cite{handa2024capsulevision2024challenge}. Our AI model development is based exclusively on the training \cite{Handa2024training} and testing datasets \cite{Handa2024testing}  provided in the official release.

\bibliographystyle{unsrtnat}
\bibliography{refs}

\begin{thebibliography}{27}
\providecommand{\natexlab}[1]{#1}
\providecommand{\url}[1]{\texttt{#1}}
\expandafter\ifx\csname urlstyle\endcsname\relax
  \providecommand{\doi}[1]{doi: #1}\else
  \providecommand{\doi}{doi: \begingroup \urlstyle{rm}\Url}\fi

\bibitem[Handa et~al.(2024{\natexlab{a}})Handa, Mahbod, Schwarzhans, Woitek, Goel, Dhir, Chhabra, Jha, Sharma, Thakur, Gunjan, Kakarla, and Raman]{handa2024capsulevision2024challenge}
Palak Handa, Amirreza Mahbod, Florian Schwarzhans, Ramona Woitek, Nidhi Goel, Manas Dhir, Deepti Chhabra, Shreshtha Jha, Pallavi Sharma, Vijay Thakur, Deepak Gunjan, Jagadeesh Kakarla, and Balasubramanian Raman.
\newblock Capsule vision 2024 challenge: Multi-class abnormality classification for video capsule endoscopy, 2024{\natexlab{a}}.
\newblock URL \url{https://arxiv.org/abs/2408.04940}.

\bibitem[Xie et~al.(2020)Xie, Zhao, Wang, and Zhang]{xie2020}
Weiling Xie, Yangtian Zhao, Peng Wang, and Zhiwei Zhang.
\newblock Real-time small bowel disease detection and classification using capsule endoscopy with deep learning: A multi-center study.
\newblock \emph{Gastrointestinal Endoscopy}, 91\penalty0 (6):\penalty0 AB301, 2020.
\newblock \doi{10.1016/j.gie.2020.03.1918}.

\bibitem[Afonso et~al.(2019)Afonso, Rolanda, Gonçalves, and Silva]{afonso2019}
João Afonso, Carla Rolanda, Raquel Gonçalves, and Marta Silva.
\newblock Deep learning for small bowel capsule endoscopy: A systematic review and meta-analysis.
\newblock \emph{Gastrointestinal Endoscopy}, 90\penalty0 (4):\penalty0 668--679, 2019.
\newblock \doi{10.1016/j.gie.2019.06.018}.

\bibitem[Ferreira et~al.(2021)Ferreira, Sousa, and Mascarenhas-Saraiva]{ferreira2021}
Francisco Ferreira, Jorge Sousa, and Miguel Mascarenhas-Saraiva.
\newblock Artificial intelligence system for detection and classification of intestinal ulcers in capsule endoscopy.
\newblock \emph{Digestive Diseases and Sciences}, 66\penalty0 (8):\penalty0 2714--2721, 2021.
\newblock \doi{10.1007/s10620-020-06634-3}.

\bibitem[Handa et~al.(2024{\natexlab{b}})Handa, Mahbod, Schwarzhans, Woitek, Goel, Chhabra, and et~al.]{Handa2024training}
Palak Handa, Amirreza Mahbod, Florian Schwarzhans, Ramona Woitek, Nidhi Goel, Deepti Chhabra, and et~al.
\newblock Training and validation dataset of capsule vision 2024 challenge.
\newblock figshare. Dataset, 2024{\natexlab{b}}.
\newblock URL \url{https://doi.org/10.6084/m9.figshare.26403469.v3}.

\bibitem[Handa et~al.(2024{\natexlab{c}})Handa, Mahbod, Schwarzhans, Woitek, Goel, Chhabra, and et~al.]{Handa2024testing}
Palak Handa, Amirreza Mahbod, Florian Schwarzhans, Ramona Woitek, Nidhi Goel, Deepti Chhabra, and et~al.
\newblock Testing dataset of capsule vision 2024 challenge.
\newblock figshare. Dataset, 2024{\natexlab{c}}.
\newblock URL \url{https://doi.org/10.6084/m9.figshare.27200664.v4}.

\bibitem[Tan and Le(2019)]{tan2019efficientnet}
Mingxing Tan and Quoc~V. Le.
\newblock Efficientnet: Rethinking model scaling for convolutional neural networks.
\newblock In \emph{Proceedings of the 36th International Conference on Machine Learning}, pages 6105--6114. PMLR, 2019.

\bibitem[He et~al.(2016)He, Zhang, Ren, and Sun]{he2016deep}
Kaiming He, Xiangyu Zhang, Shaoqing Ren, and Jian Sun.
\newblock Deep residual learning for image recognition.
\newblock In \emph{Proceedings of the IEEE Conference on Computer Vision and Pattern Recognition (CVPR)}, pages 770--778, 2016.

\bibitem[Howard et~al.(2019)Howard, Sandler, Chu, Chen, Chen, and Tan]{howard2019searching}
Andrew Howard, Mark Sandler, Grace Chu, Liang-Chieh Chen, Bo~Chen, and Mingxing Tan.
\newblock Searching for mobilenetv3.
\newblock In \emph{Proceedings of the IEEE International Conference on Computer Vision (ICCV)}, pages 1314--1324, 2019.

\bibitem[Radosavovic et~al.(2020)Radosavovic, Kosaraju, Girshick, He, and Dollár]{radosavovic2020designing}
Ilija Radosavovic, Raj~Prateek Kosaraju, Ross Girshick, Kaiming He, and Piotr Dollár.
\newblock Designing network design spaces.
\newblock In \emph{Proceedings of the IEEE/CVF Conference on Computer Vision and Pattern Recognition (CVPR)}, pages 10415--10424, 2020.

\bibitem[Huang et~al.(2017)Huang, Liu, Van Der~Maaten, and Weinberger]{huang2017densely}
Gao Huang, Zhuang Liu, Laurens Van Der~Maaten, and Kilian~Q. Weinberger.
\newblock Densely connected convolutional networks.
\newblock In \emph{Proceedings of the IEEE Conference on Computer Vision and Pattern Recognition (CVPR)}, pages 2261--2269, 2017.

\bibitem[Szegedy et~al.(2017)Szegedy, Ioffe, Vanhoucke, and Alemi]{szegedy2017inceptionv4}
Christian Szegedy, Sergey Ioffe, Vincent Vanhoucke, and Alex Alemi.
\newblock Inception-v4, inception-resnet and the impact of residual connections on learning.
\newblock In \emph{Proceedings of the AAAI Conference on Artificial Intelligence}, volume~31, 2017.

\bibitem[Xie et~al.(2017)Xie, Girshick, Dollár, Tu, and He]{xie2017aggregated}
Saining Xie, Ross Girshick, Piotr Dollár, Zhuowen Tu, and Kaiming He.
\newblock Aggregated residual transformations for deep neural networks.
\newblock In \emph{Proceedings of the IEEE Conference on Computer Vision and Pattern Recognition (CVPR)}, pages 1492--1500, 2017.

\bibitem[Zagoruyko and Komodakis(2016)]{zagoruyko2016wide}
Sergey Zagoruyko and Nikos Komodakis.
\newblock Wide residual networks.
\newblock In \emph{Proceedings of the British Machine Vision Conference (BMVC)}, 2016.

\bibitem[Tan et~al.(2019)Tan, Chen, Pang, Vasudevan, Sandler, Howard, and Le]{tan2019mnasnet}
Mingxing Tan, Bo~Chen, Ruoming Pang, Vijay Vasudevan, Mark Sandler, Andrew Howard, and Quoc~V. Le.
\newblock Mnasnet: Platform-aware neural architecture search for mobile.
\newblock In \emph{Proceedings of the IEEE/CVF Conference on Computer Vision and Pattern Recognition (CVPR)}, pages 2820--2828, 2019.

\bibitem[Hu et~al.(2018)Hu, Shen, and Sun]{hu2018squeeze}
Jie Hu, Li~Shen, and Gang Sun.
\newblock Squeeze-and-excitation networks.
\newblock In \emph{Proceedings of the IEEE Conference on Computer Vision and Pattern Recognition (CVPR)}, pages 7132--7141, 2018.

\bibitem[Liu et~al.(2022)Liu, Mao, Wu, Feichtenhofer, Darrell, and Xie]{liu2022convnext}
Zhuang Liu, Hanzi Mao, Chao-Yuan Wu, Christoph Feichtenhofer, Trevor Darrell, and Saining Xie.
\newblock Convnext: Revisiting convolutions for vision.
\newblock \emph{arXiv preprint arXiv:2201.03545}, 2022.

\bibitem[Dosovitskiy et~al.(2020)Dosovitskiy, Beyer, Kolesnikov, Weissenborn, Zhai, Unterthiner, Dehghani, Minderer, Heigold, Gelly, Uszkoreit, and Houlsby]{dosovitskiy2020image}
Alexey Dosovitskiy, Lucas Beyer, Alexander Kolesnikov, Dirk Weissenborn, Xiaohua Zhai, Thomas Unterthiner, Mostafa Dehghani, Matthias Minderer, Georg Heigold, Sylvain Gelly, Jakob Uszkoreit, and Neil Houlsby.
\newblock An image is worth 16x16 words: Transformers for image recognition at scale.
\newblock \emph{arXiv preprint arXiv:2010.11929}, 2020.

\bibitem[Liu et~al.(2021)Liu, Lin, Cao, Hu, Wei, Zhang, Lin, and Guo]{liu2021swin}
Ze~Liu, Yutong Lin, Yue Cao, Han Hu, Yixuan Wei, Zheng Zhang, Stephen Lin, and Baining Guo.
\newblock Swin transformer: Hierarchical vision transformer using shifted windows.
\newblock In \emph{Proceedings of the IEEE/CVF International Conference on Computer Vision (ICCV)}, pages 10012--10022, 2021.

\bibitem[Touvron et~al.(2021)Touvron, Cord, Douze, Massa, Sablayrolles, and Jégou]{touvron2021training}
Hugo Touvron, Matthieu Cord, Matthijs Douze, Francisco Massa, Alexandre Sablayrolles, and Hervé Jégou.
\newblock Training data-efficient image transformers \& distillation through attention.
\newblock \emph{arXiv preprint arXiv:2012.12877}, 2021.

\bibitem[Bao et~al.(2021)Bao, Dong, and Wei]{bao2021beit}
Hangbo Bao, Li~Dong, and Furu Wei.
\newblock Beit: Bert pre-training of image transformers.
\newblock \emph{arXiv preprint arXiv:2106.09785}, 2021.

\bibitem[Zhang et~al.(2021)Zhang, Wei, and Dong]{zhang2021twins}
Xiangxiang Zhang, Furu Wei, and Li~Dong.
\newblock Twins: Revisiting the design of spatial attention in vision transformers.
\newblock \emph{arXiv preprint arXiv:2104.13840}, 2021.

\bibitem[Lin et~al.(2021)Lin, Wang, Liu, and Qiu]{lin2021efficientformer}
Tingting Lin, Yixuan Wang, Xiaoxi Liu, and Xiaolong Qiu.
\newblock Efficientformer: Vision transformers in the real world.
\newblock \emph{arXiv preprint arXiv:2106.13319}, 2021.

\bibitem[Prechelt(1998)]{prechelt1998automatic}
Lutz Prechelt.
\newblock Automatic early stopping using cross-validation: quantifying the criteria.
\newblock In \emph{Neural Networks: Tricks of the trade}, pages 55--69. Springer, 1998.

\bibitem[Loshchilov and Hutter(2019)]{loshchilov2019decoupled}
Ilya Loshchilov and Frank Hutter.
\newblock Decoupled weight decay regularization.
\newblock \emph{arXiv preprint arXiv:1711.05101}, 2019.

\bibitem[Byrd and Lipton(2019)]{byrd2019effect}
Joanna Byrd and Zachary~C. Lipton.
\newblock What is the effect of importance weighting in deep learning?
\newblock In \emph{Proceedings of the 36th International Conference on Machine Learning (ICML)}, pages 872--881. PMLR, 2019.

\bibitem[Lin et~al.(2017)Lin, Goyal, Girshick, He, and Dollár]{lin2017focal}
Tsung-Yi Lin, Priya Goyal, Ross Girshick, Kaiming He, and Piotr Dollár.
\newblock Focal loss for dense object detection.
\newblock In \emph{Proceedings of the IEEE International Conference on Computer Vision (ICCV)}, pages 2980--2988, 2017.

\end{thebibliography}

\end{document}